\documentclass[10pt,twocolumn,letterpaper]{article}
\usepackage{iccv}
\usepackage{times}
\usepackage{epsfig}
\usepackage{graphicx}
\usepackage{amsmath}
\usepackage{amssymb}
\usepackage{multirow}
\usepackage{sidecap}
\usepackage{bm}
\usepackage{textgreek}
\usepackage{subfig}
\usepackage[table]{xcolor}
\usepackage{diagbox}
\usepackage{algcompatible}
\usepackage[linesnumbered,ruled,vlined]{algorithm2e}

\usepackage{sidecap}

\usepackage[pagebackref=true,breaklinks=true,letterpaper=true,colorlinks,bookmarks=false]{hyperref}

\newcommand{\bx}{{\bm x}}
\newcommand{\btheta}{{\bm \theta}}
\newcommand{\bdelta}{{\bm \delta}}

\iccvfinalcopy 

\def\iccvPaperID{2247} 

\ificcvfinal\fi
\begin{document}

\title{\vspace{-1cm} Adversarial Defense by Restricting the Hidden Space\\ of Deep Neural Networks}

\vspace{-1cm}
\author{Aamir Mustafa$^{1,3}$ \qquad Salman Khan$^{1,2}$ \qquad Munawar Hayat$^{1,3}$ \\
Roland Goecke$^{3}$\qquad Jianbing Shen$^{1,4}$ \qquad Ling Shao$^{1}$\\
$^1$Inception Institute of Artificial Intelligence, $^2$Australian National University,\\
$^3$University of Canberra, $^4$Beijing Institute of Technology\\
}

\maketitle

\begin{abstract}
Deep neural networks are vulnerable to adversarial attacks, which can fool them by adding minuscule perturbations to the input images. The robustness of existing defenses suffers greatly under white-box attack settings, where an adversary has full knowledge about the network and can iterate several times to find strong perturbations. We observe that the main reason for the existence of such perturbations is the close proximity of different class samples in the learned feature space. This allows model decisions to be totally changed by adding an imperceptible perturbation in the inputs. To counter this, we propose to class-wise disentangle the intermediate feature representations of deep networks. Specifically, we force the features for each class to lie inside a convex polytope that is maximally separated from the polytopes of other classes. In this manner, the network is forced to learn distinct and distant decision regions for each class. We observe that this simple constraint on the features greatly enhances the robustness of learned models, even against the strongest white-box attacks, without degrading the classification performance on clean images. We report extensive evaluations in both black-box and white-box attack scenarios and show significant gains in comparison to state-of-the art defenses\footnote{Code and models are available at: \url{https://github.com/aamir-mustafa/pcl-adversarial-defense}}. 
\end{abstract}

\vspace{-1em}
\section{Introduction}
\label{sec:intro}
Adversarial examples contain small, human-imperceptible perturbations specifically designed by an adversary to fool a learned model \cite{szegedy2013intriguing, 43405}. These examples pose a serious threat for security critical applications, \eg autonomous cars \cite{ackerman2017drive}, bio-metric identification \cite{sanderson2008biometric} and surveillance systems \cite{najafabadi2015deep}. Furthermore, if a slight perturbation added to a benign input drastically changes the deep network's output with a high-confidence, it reflects that our current models are not distinctively learning the fundamental visual concepts. Therefore, the design of robust deep networks goes a long way towards developing reliable and trustworthy artificial intelligence systems. 

\begin{figure}[t]
    \centering
   {\includegraphics[trim={0cm 4.9cm 13.7cm 0cm}, clip, width=0.5\textwidth]{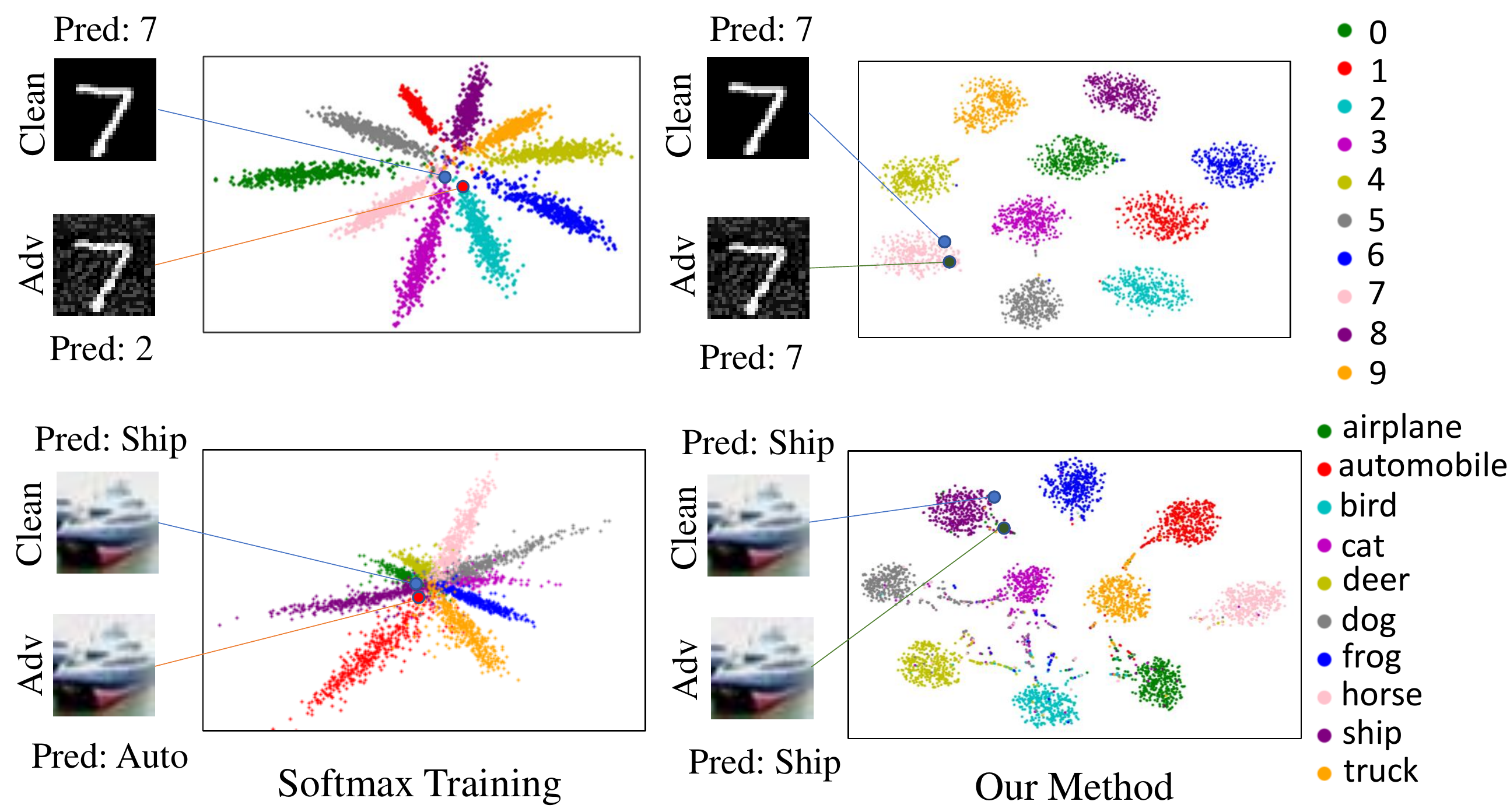}}
    \caption{\small{2D penultimate layer activations of a clean image and its adversarial counterpart (PGD attack) for standard softmax trained model and our method on MNIST (top row) and CIFAR-10 (bottom row) datasets. Note that our method correctly maps the attacked image to its true-class feature space.}} \vspace{-1.5em}
    \label{fig:comparison}%
\end{figure}


To mitigate adversarial attacks, various defense methods have recently been proposed. These can be broadly classified into two categories: (a) \textit{Reactive defenses} that modify the inputs during testing time, using image transformations to counter the effect of adversarial perturbation \cite{luo2015foveation, das2017keeping, xie2017mitigating, mustafa2019image}, and (b) \emph{Proactive defenses} that alter the underlying architecture or learning procedure \eg by adding more layers, ensemble/adversarial training or changing the loss/activation functions \cite{tramer2017ensemble, kurakin2016adversarial, papernot2016distillation, cisse2017parseval, liao2017defense, pang2019improving, madry2017towards, kannan2018adversarial}. Proactive defenses are generally more valued, as they provide relatively better robustness against \textit{white-box} attacks. Nevertheless, both proactive and reactive defenses are easily circumvented by the iterative \textit{white-box} adversaries \cite{athalye2018obfuscated}.

This paper introduces a new proactive defense based on a novel training procedure, which maximally separates the learned feature representations at multiple depth levels of the deep model. We note that the addition of perturbations in the input domain leads to a corresponding polytope in the high-dimensional manifold of the intermediate features and the output classification space. Based upon this observation, we propose to maximally separate the polytopes for different class samples, such that there is a minimal overlap between any two classes in the decision and intermediate feature space. This ensures that an adversary can no longer fool the network within a restricted perturbation budget. In other words, we build on the intuition that two different class samples, which are visually dissimilar in the input domain, must be mapped to different regions in the output space. Therefore, we must also enforce that their feature representations are well separated along the hierarchy of network layers. This is achieved by improving within-class proximities and enhancing between-class differences of the activation maps, along multiple levels of the deep model. As illustrated in Fig.~\ref{fig:comparison}, the penultimate layer features learnt by the proposed scheme are well separated and hard to penetrate compared with the easily attacked features learnt using standard loss without any deep supervision. As evidenced with empirical evaluations (Sec.~\ref{sec:experiments}), the proposed method provides an effective and robust defense by significantly outperforming current state-of-the-art defenses under both \textit{white-box} and \textit{black-box} settings. Also, we experimentally show that our method does not suffer from the obfuscated gradient problem, which is otherwise the case for most existing defenses.


Our approach provides strong evidence towards the notion that the adversarial perturbations exist not only due to the properties of data (\eg high-dimensionality) and network architecture (\eg non-linearity functions) but also are greatly influenced by the choice of objective functions used for optimization. The deeply supervised multi-layered loss based defense provides a significant boost in robustness under strictest attack conditions where the balance is shifted heavily towards the adversary. These include \emph{white-box} attacks and iterative adversaries including the strongest first-order attacks (Projected Gradient Descent). We demonstrate the robustness of the proposed defense through extensive evaluations on five publicly available datasets and achieve a robustness of $46.7\%$ and $36.1\%$ against the strongest PGD attack ($\epsilon = 0.03$) for the CIFAR-10 and CIFAR-100 datasets, respectively. To the best of our knowledge, these are significantly higher levels of robustness against a broad range of strong adversarial attacks.

\section{Related Work}
\label{sec:related-work}

Generating adversarial examples to fool a deep network and developing defenses against such examples have gained significant research attention recently. Adversarial perturbations were first proposed by Szegedy \textit{et al.}\ \cite{szegedy2013intriguing} using an L-BFGS based optimization scheme, followed by Fast Gradient Sign Method (FGSM) \cite{43405} and its iterative variant \cite{kurakin2016adversarial}. Moosavi-Dezfooli \textit{et al.}\ \cite{moosavi2016deepfool} then proposed DeepFool, which iteratively projects an image across the decision boundary (form of a polyhydron) until it crosses the boundary and is mis-classified. One of the strongest attacks proposed recently is the Projected Gradient Descent (PGD) \cite{madry2017towards}, which takes maximum loss increments allowed within a specified $l_\infty$ norm-ball. Other popular attacks include the Carlini and Wagner Attack \cite{carlini2017towards}, Jacobian-based Saliency Map Approach \cite{papernot2016limitations}, Momentum Iterative Attack \cite{dong2018boosting} and Diverse Input Iterative Attack \cite{xie2018improving}.

Two main lines of defense mechanisms have been proposed in the literature to counter adversarial attacks. First, by applying different pre-processing steps and transformations on the input image at inference time \cite{xie2017mitigating,guo2017countering}. The second category of defenses improve network's training regime to counter adversarial attacks. An effective scheme in this regards is \textit{adversarial training}, where the model is jointly trained with clean images and their adversarial counterparts \cite{kurakin2016adversarial_2, 43405}. Ensemble adversarial training is used in \cite{tramer2017ensemble} to soften the classifier's decision boundaries. Virtual Adversarial Training \cite{miyato2015distributional} smoothes the model distribution using a regularization term. Papernot \textit{et al.} \cite{papernot2016distillation} used distillation to improve the model's robustness by retraining with soft labels. Parsevel Networks \cite{cisse2017parseval} restrict the Lipschitz constant of each layer of the model. Input Gradient Regularizer \cite{ross2018improving} penalizes the change in model's prediction w.r.t input perturbations by regularizing the gradient of cross-entropy loss. The Frobenius norm of the Jacobian of the network has been shown to improve model's stability in \cite{jakubovitz2018improving}. \cite{lin2018defensive} proposed defensive quantization method to control the Lipschitz constant of the network to mitigate the adversarial noise during inference. \cite{dhillon2018stochastic} proposed Stochastic Activation Pruning as a defense against adversarial attacks. Currently the strongest defense method is Min-Max optimization \cite{madry2017towards} which augments the training data with a first order attacked samples. Despite significant research activity in devising defenses against adversarial attacks, it was recently shown in \cite{athalye2018obfuscated} that the currently existing state-of-the-art defenses \cite{kolter2017provable,raghunathan2018certified,sinha2017certifiable} are successfully circumvented under \textit{white-box} settings. Only Min-Max optimization \cite{madry2017towards} and Cascade adversarial machine learning \cite{na2017cascade} retained $47\%$ and $15\%$ accuracy respectively, and withstood the attacks under \textit{white-box} settings. In our experiments (see Sec.~\ref{sec:experiments}), we extensively compare our results with \cite{madry2017towards} and make a compelling case by achieving significant improvements.


At the core of our defense are the proposed objective function and multi-level deep supervision, which ensure feature space discrimination between classes. Our training objective is inspired from center loss \cite{wen2016discriminative}, which clusters penultimate layer features. We propose multiple novel constraints (Sec.~\ref{sec:background}) to enhance between-class distances, and ensure maximal separation of a sample from its non-true classes. Our method is therefore fundamentally different from \cite{wen2016discriminative}, since the proposed multi-layered hierarchical loss formulation and the notion of maximal separation has not been previously explored for adversarial robustness. 

\begin{figure}
    \centering
   {\includegraphics[trim={2.5cm 6.55cm 4.3cm 0cm}, clip, width=1\linewidth]{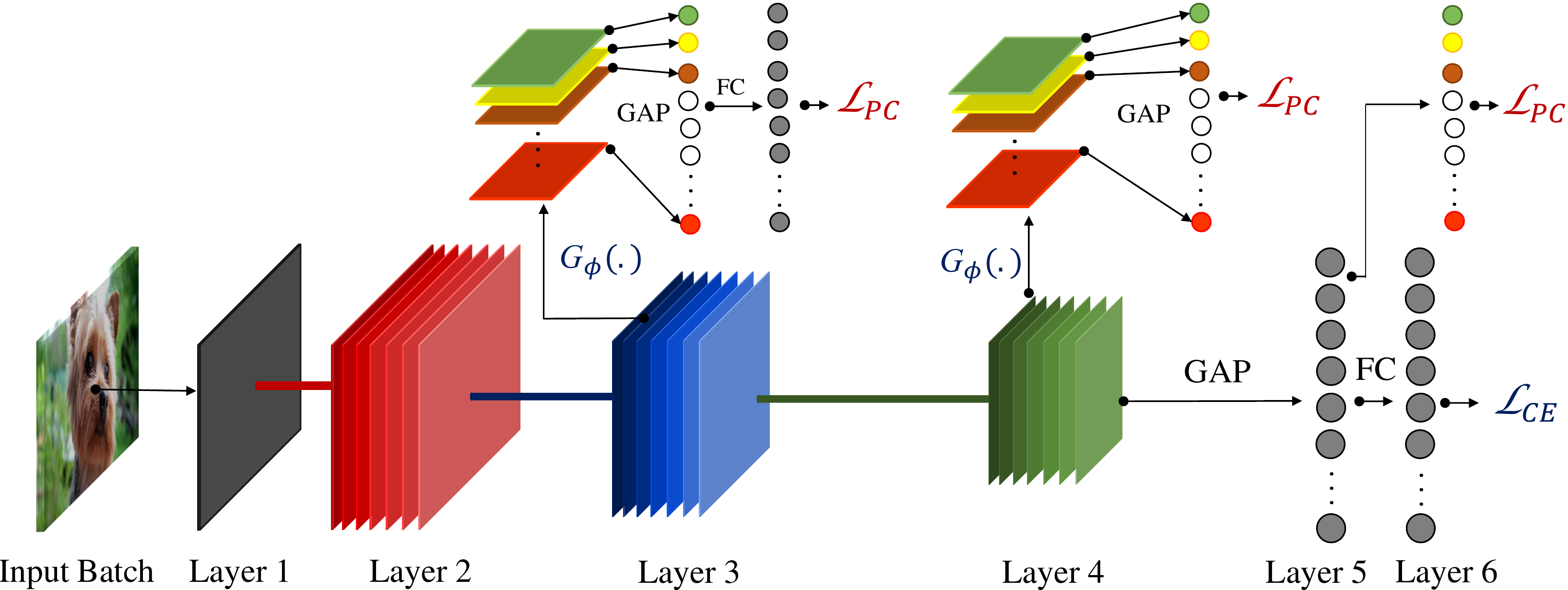} }
   \vspace{-1em}
    \caption{\small{An illustration of our training with joint supervision of  $\mathcal{L}_{\text{PC}}$ and  $\mathcal{L}_{\text{CE}}$. $G_{\phi}(\cdot)$ is an auxiliary branch to map features to a low dimensional output, which is then used for loss in Eq.~\ref{eq:auxiliary}}}
    \label{fig:block_diag}%
    \vspace{-0.5em}    
\end{figure}

\section{Prototype Conformity Loss}
\label{sec:background}

Below, we first introduce the notations used, then provide a brief overview of the conventional cross entropy loss followed by a detailed description of our proposed method.

\noindent \textbf{Notations:} Let $\bm{x} \in \mathbb{R}^{m}$ and $\bm{y}$ denote an input-label pair and $\bm{1}_{y}$ be the one-hot encoding of $\bm{y}$. We denote a deep neural network (DNN) as a function $\mathcal{F}_{\bm{\theta}}(\bm{x})$,  where $\bm{\theta}$ are the trainable parameters. The DNN outputs a feature representation $\bm{f}\in \mathbb{R}^d$, which is then used by a classification layer to perform multi-class classification. Let $k$ be the number of classes; the parameters of the classifier can then be represented as $\bm{W} = [\bm{w}_1,\ldots,\bm{w}_k] \in \mathbb{R}^{d\times k}$. To train the model, we find the optimal $\bm{\theta}$ and $\bm{W}$ that minimize a given objective function. Next, we introduce a popular loss function for deep CNNs.

\noindent \textbf{Cross-entropy Objective:}  The cross-entropy objective function maximizes the dot product between an input feature $\bm{f}_i$ and its true class representative vector $\bm{w}_y$, such that $\bm{w}_y \in \bm{W}$. In other words, cross-entropy loss forces the classifier to learn a mapping from feature to output space such that the projection on to the correct class vector is maximized:\vspace{-0.55em}
\begin{equation}
\label{eq:2}
\mathcal{L}_{\text{CE}}(\bm{x}, \bm{y})= \sum_{i=1}^m - \log\frac{\exp({\bm{w}^T_{y_i}\bm{f}_i + \bm{b}_{y_i}) }}{\sum_{j=1}^k \exp({\bm{w}^T_{j}\bm{f}_i + \bm{b}_{j}}) },
\end{equation}
where, $m$ is the number of images, and $\bm{f}_i$ is the feature of an $i^{th}$ image $\bm{x}_i$ with the class $\bm{y}_i$. $\bm{W}$ and $\bm{b}$ are, respectively, the weights and the bias terms for the classification layer.

\noindent \textbf{Adversarial Perspective:} The main goal of an attack algorithm is to force a trained DNN $\mathcal{F}_\theta$ to make wrong predictions. Attack algorithms seek to achieve this goal within a minimal perturbation budget. The attacker's objective can be represented by:\vspace{-0.55em}
\begin{equation}
    \label{eq:adv_loss}
    \underset{\bdelta}{\text{argmax}}\, \mathcal{L}(\bx + \bdelta,\bm{y}), \quad s.t., {\parallel}\bdelta{\parallel}_p < \epsilon,
    \vspace{-0.5em}
\end{equation}
where $\bm{y}$ is the ground-truth label for an input sample $\bx$,  $\bdelta$ denotes the adversarial perturbation, $\mathcal{L}(\cdot)$ denotes the error function, $\parallel\cdot\parallel_p$ denotes the p-norm, which is generally considered to be an $\ell_{\infty}$-ball centered at $\bm{x}$, and $\epsilon$ is the available perturbation budget.

In order to create a robust model, the learning algorithm must consider the allowed perturbations in the input domain and learn a function that maps the perturbed images to the correct class. This can be achieved through the following min-max (saddle point) objective that minimizes the empirical risk in the presence of perturbations:
\begin{align}\label{eq:adv_training}
    \min_{\theta} \underset{(\bx,\bm{y}) \sim \mathcal{D}}{\mathbb{E}} \big[\max_{\delta} \mathcal{L}(\bm{x}+ \bdelta, \bm{y}; \bm{\theta})\big], \; s.t., {\parallel}\bdelta{\parallel}_p < \epsilon,
\end{align}
where $\mathcal{D}$ is the data distribution.

\noindent \textbf{CE Loss in Adversarial Setting:} The CE loss is the default choice for conventional classification tasks. However, it simply assigns an input sample to one of the pre-defined classes. It therefore does not allow one to distinguish between normal and abnormal inputs (adversarial perturbations in our case). Further, it does not explicitly enforce any margin constraints amongst the learned classification regions. It can be seen from Eq.~\ref{eq:adv_training} that an adversary's job is to maximize $\mathcal{L}(\cdot)$ within a small perturbation budget $\epsilon$. Suppose, the adversarial polytope in the output space\footnote{Note that the output space in our case is not the final prediction space, but the intermediate feature space.} with respect to an input sample $\bm{x}$ is given by:
\begin{align}
    \mathcal{P}_{\epsilon}(\bx; \theta) = \{\mathcal{F}_{\theta}(\bx +\bdelta) \;s.t., \, {\parallel}\bdelta{\parallel}_p \leq \epsilon\}.
\end{align}
An adversary's task is easier if there is an overlap between the adversarial polytopes for different input samples belonging to different classes. 

\noindent
\textbf{Definition 1:} \emph{The overlap $\mathcal{O}_{\epsilon}^{i,j}$ between polytopes for each data sample pair $(i,j)$ can be defined as the volume of intersection between the respective polytopes:}
$$ \mathcal{O}_{\epsilon}^{i,j} = \mathcal{P}_{\epsilon}(\bx^i_{y_i}; \theta) \cap \mathcal{P}_{\epsilon}(\bx^j_{y_j}; \theta).$$

\noindent Note that the considered polytopes can be non-convex as well. However, the overlap computation can be simplified for convex polytopes \cite{de1998computing}. 

\noindent
\textbf{Proposition 1:} \emph{For an $i^{th}$ input sample $\bx^{i}_{y_i}$ with class label $\bm{y}_i$, reducing the overlap $\mathcal{O}_{\epsilon}^{i,j}$ between its polytope $\mathcal{P}_{\epsilon}(\bx^i_{y_i}; \theta)$ and the polytopes of other class samples $\mathcal{P}_{\epsilon}(\bx^j_{y_j}; \theta)$, s.t., $\bm{y}_j\neq \bm{y}_i$ will result in lower adversary success for a bounded perturbation ${\parallel}\bdelta{\parallel}_p \leq \epsilon$.}

\noindent
\textbf{Proposition 2:} \emph{For a given adversarial strength $\epsilon$, assume $\lambda$ is the maximum distance from the center of the polytope to the convex outer bounded polytope. Then, a classifier maintaining a margin $m>2\lambda$ between two closest samples belonging to different classes will result in a decision boundary with guaranteed robustness against perturbation within the budget $\epsilon$.}

In other words, if the adversarial polytopes for samples belonging to different classes are non-overlapping, the adversary cannot find a viable perturbation within the allowed budget. We propose that an adversary's task can be made difficult by including a simple maximal separation constraint in the objective of deep networks. The conventional CE loss does not impose any such constraint, which makes the resulting models weaker against adversaries.
A more principled approach is to define convex category-specific classification regions for each class, where any sample outside all of such regions is considered an adversarial perturbation. Consequently, we propose the prototype conformity loss function, described below. \\

\vspace{-0.5em}
\noindent \textbf{Proposed Objective:} We represent each class with its prototype vector, which represents the training examples of that class.  Each class is assigned a fixed and non-overlapping p-norm ball and the training samples belonging to a class $i$ are encouraged to be mapped close to its hyper-ball center:
\vspace{-0.55em}
\begin{align}
\label{eq:5}
    \mathcal{L}_{\text{PC}}(\bm{x}, \bm{y}) = & \,\sum\limits_{i}\bigg\{ {\parallel}\bm{f}_i - \bm{w}_{y_i}^c{\parallel}_2 - \frac{1}{k-1} \sum\limits_{j\neq y_i}\Big( {\parallel} \bm{f}_i - \bm{w}_j^c{\parallel}_2 \notag\\ 
    & +  {\parallel} \bm{w}_{y_i}^c - \bm{w}_{j}^c{\parallel}_2 \Big) \bigg\}.
\end{align}
During model inference, a feature's similarity is computed with all the class prototypes and it is assigned the closest class label if and only if the sample lies within its decision region: 
\vspace{-0.95em}
\begin{align}
  \hat{\bm{y}}_i = \underset{j}{\text{argmin}} \,{\parallel}\bm{f}_i - \bm{w}_j^c {\parallel}.
\end{align}

Here, $\bm{w}^c$ denotes the trainable class centroids. Note that the classification rule is similar to the Nearest Class Mean (NCM) classifier \cite{mensink2013distance}, but we differ in some important aspects: (a) the centroids for each class are not fixed as the mean of training samples, rather learned automatically during representation learning, (b) class samples are explicitly forced to lie within respective class norm-balls, (c) feature representations are appropriately tuned to learn discriminant mappings in an end-to-end manner, and (d) to avoid inter-class confusions, disjoint classification regions are considered by maintaining a large distance between each pair of prototypes. We also experiment with the standard softmax classifier and get equivalent performance compared to nearest prototype rule mentioned above.

\noindent \textbf{Deeply Supervised Learning:} 
The overall loss function used for training our model is given by:
\begin{align}
    \mathcal{L}(\bm{x},\bm{y}) = \mathcal{L}_{\text{CE}}(\bm{x},\bm{y}) +  \mathcal{L}_{\text{PC}}(\bm{x},\bm{y}).
\end{align}
The above loss enforces the intra-class compactness and an inter-class separation using learned prototypes in the output space. In order to achieve a similar effect in the intermediate feature representations, we include other auxiliary loss functions $\{ \mathcal{L}^n\}$ along the depth of our deep networks, which act as companion objective functions for the final loss. This is achieved by adding an auxiliary branch $\mathcal{G}_{\phi}(\cdot)$ after the defined network depth, which maps the features to a lower dimension output, and is then used in the loss definition. For illustration, see Fig.~\ref{fig:block_diag}.
\begin{align}
    \label{eq:auxiliary}
    \mathcal{L}^n(\bm{x}, \bm{y}) &= \mathcal{L}_{\text{CE}}(\bm{f}^l, \bm{y}) + \mathcal{L}_{\text{PC}}(\bm{f}^l,\bm{y})\\
    &s.t., \; \bm{f}^l = \mathcal{G}_{\phi}^l(\mathcal{F}^l_{\theta}(\bm{x})).  
\end{align}
These functions avoid the vanishing gradients problem and act as regularizers that encourage features belonging to the same class to come together and the ones belonging to different classes to be pushed apart. 

\begin{figure}[t] 
    \centering
   {\includegraphics[trim={0cm 7.35cm 4.8cm 0cm}, clip, width=0.5\textwidth]{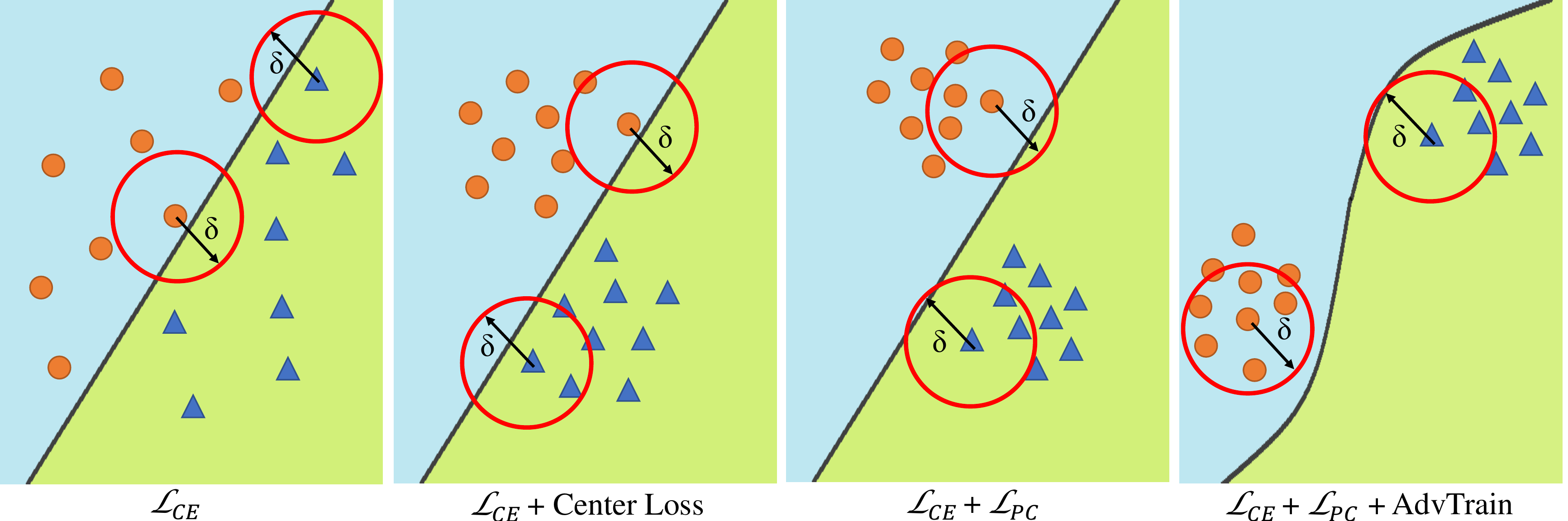}}\vspace{-0.5em}
    \caption{\small{Comparison between different training methods. The red circle encompasses the adversarial sample space within a perturbation budget ${\parallel}\bdelta{\parallel}_p < \epsilon$.}} \vspace{-1.25em}
    \label{fig:toy-fig}
\end{figure}

\section{Adversarial Attacks}
\label{subsec:attacks}

We evaluate our defense model against five recently proposed state-of-the-art attacks, which are summarized below, for completeness.\medskip  \\
\vspace{-1.0em}

\noindent \textbf{Fast Gradient Sign Method (FGSM)} \cite{43405} generates an adversarial sample $\bm{x}_{adv}$ from a clean sample $\bm{x}$ by maximizing the loss in Eq.~\ref{eq:adv_loss}. It finds $\bm{x}_{adv}$ by moving a single step in the opposite direction to the gradient of the loss function, as:
\vspace{-0.7em}
\begin{equation}
    \label{eq:fgsm}
    \bm{x}_{adv} = \bm{x} + \epsilon \cdot \text{sign}(\nabla_x \mathcal{L}(\bx,\bm{y})).
\end{equation}
Here, $\epsilon$ is the allowed perturbation budget. \medskip \\ 
\textbf{Basic Iterative Method (BIM)} \cite{kurakin2016adversarial} is an iterative variant of FGSM and generates an adversarial sample as:
\vspace{-0.75em}
\begin{equation}
\label{eq-ifgsm}
 \bm{x}_{m} = \text{clip}_{\epsilon} (\bm{x}_{m-1} + \frac{\epsilon}{i} \cdot \text{sign}(\nabla_{x_{m-1}} (\mathcal{L}(\bm{x}_{m-1},\bm{y}))),
\end{equation}
where $\bm{x}_0$ is clean image $\bm{x}$ and \textit{i} is the iteration number. \medskip \\ 
\textbf{Momentum Iterative Method (MIM)} \cite{dong2018boosting} introduces an additional momentum term to BIM to stabilize the direction of gradient. Eq.\ \ref{eq-ifgsm} is modified as:
\vspace{-0.75em}
\begin{equation}
\label{eq:mifgsm}
 g_{m} = \mu \cdot g_{m-1} + \frac{\nabla_{x_{m-1}} \mathcal{L}(\bm{x}_{m-1},\bm{y})}{\parallel \nabla_{x_{m-1}} (\mathcal{L}(\bm{x}_{m-1},\bm{y})) \parallel_1}
\end{equation}
\vspace{-0.75em}
\begin{equation}
 \bm{x}_{m} = \text{clip}_{\epsilon} (\bm{x}_{m-1} + \frac{\epsilon}{i} \cdot \text{sign}(g_{m})),
\end{equation}
where $\mu$ is the decay factor. \medskip \\
\textbf{Carlini \& Wagner Attack} \cite{carlini2017towards} defines an auxiliary variable $\zeta$ and minimizes the objective function:
\vspace{-0.75em}
\begin{equation}
\underset{\zeta}{\text{min}} \parallel \frac{1}{2}(\tanh{(\zeta)}+1) - \bm{x} \parallel + c \cdot f(\frac{1}{2}(\tanh{\zeta}+1)),
\vspace{-0.5em}
\end{equation}
where $ \frac{1}{2}(\tanh{(\zeta)}+1) - \bm{x} $ is the perturbation $\bdelta$, $c$ is the constant chosen and $f(.)$ is defined as:
\vspace{-0.25em}
\begin{equation}
f(\bm{x}_{adv})= \max(\mathcal{Z}(\bm{x}_{adv})_{\bm{y}} - \max \{\mathcal{Z}(\bm{x}_{adv})_k : k \neq \bm{y} \}, -\kappa).
\end{equation}
Here, $\kappa$ controls the adversarial sample's confidence and $\mathcal{Z}(\bm{x}_{adv})_k$ are the logits values corresponding to a class $k$.\medskip \\
\textbf{Projected Gradient Descent (PGD)} \cite{madry2017towards} is similar to BIM, and starts from a random position in the clean image neighborhood $\mathcal{U}(\bx, \epsilon)$. This method applies FGSM for $m$ iterations with a step size of $\gamma$ as:
\vspace{-0.57em}
\begin{align}
    \bm{x}_{m} = & \bm{x}_{m-1} + \gamma \cdot \text{sign}(\nabla_{x_{m-1}} \mathcal{L}(\bx_{m-1},\bm{y})). \\
    & \bm{x}_{m} = \text{clip}(\bm{x}_{m}, \bm{x}_{m}-\epsilon, \bm{x}_{m} + \epsilon). 
\end{align}
It proves to be a strong iterative attack, relying on the first order information of the target model.

\vspace{-0.25em}
\section{Experiments}
\label{sec:experiments}
\vspace{-0.25em}
\noindent \textbf{Datasets and Models:} We extensively evaluate the proposed method on five datasets: MNIST, Fasion-MNIST (F-MNIST), CIFAR-10, CIFAR-100 and Street-View House Numbers (SVHN). For the MNIST and F-MNIST datasets, the CNN model chosen has six layers, as in \cite{wen2016discriminative}. For the CIFAR-10, CIFAR-100 and SVHN datasets, we use a ResNet-110 model \cite{he2016deep} (see Table~\ref{table:models}). The deep features for the prototype conformity loss are extracted from different intermediate layers using an auxiliary branch, which maps the features to a lower dimension output (see Fig.~\ref{fig:block_diag}). We first train for $T^{'}$ epochs ($T^{'}=50$ for F/MNIST, $T^{'}=200$ for CIFAR-10/100 and SVHN) with $\mathcal{L}_\text{CE}$ and then use the loss in Eq.~\ref{eq:auxiliary} for 300 epochs. A batch size of $256$ and a learning rate of $0.1$ ($\times 0.1$ at $T$=200, 250) are used. Further training details are summarized in Algorithm~\ref{algo}.

\begin{algorithm}[h]
\footnotesize{\KwIn{Classifier $\mathcal{F}_\theta(\bm{x})$, training data \{$\bx$\}, ground truth labels \{$\bm{y}$\}, trainable parameters $\bm{\theta}$, trainable class centroids \{$\bm{w}^c_{j}: j \in [1,k]$ \}, perturbation budget $\epsilon$, epochs $T$, number of auxiliary branches $L$. }}
\footnotesize{\KwOut{Updated parameters $\btheta$}}
\vspace{2mm}
Initialize $\btheta$ in convolution layers.\\
\textbf{for} $t = 0$ to $T$:\\
\qquad \textbf{if} $t < T^{'}$:\\
\qquad\qquad  Converge softmax objective, $\btheta:= \arg\min_{\btheta} \mathcal{L}_\text{CE}$.\\
\qquad \textbf{else}:\\
\qquad\qquad Compute joint loss $\mathcal{L} = \mathcal{L}_\text{CE} + \sum_{l}^{L} \mathcal{L}_\text{PC}$\\
\qquad\qquad   Compute gradients w.r.t $\btheta$ and $\bx$,  $\nabla_\theta \mathcal{L}(\bx,\bm{y})$ and $\nabla_x \mathcal{L}(\bx,\bm{y})$ respectively.\\
\qquad\qquad  Update model weights, $\btheta:= \arg\min_{\btheta} \mathcal{L}$. \\
\qquad\qquad Update class centroids $\bm{w}^c_j \; \forall \; l$ \\
\qquad\qquad  Generate adversarial examples as:\\
\qquad \qquad\qquad  \textbf{if} FGSM: \textbf{then}  $\bm{x}_{adv} = \bm{x} + \epsilon \cdot \text{sign}\,(\nabla_x \mathcal{L}(\bx,\bm{y}))$ \\
\qquad\qquad \qquad  \textbf{elif} PGD: \textbf{then} $\bm{x}_{adv} = \text{clip}\, (\bm{x}, \bx - \epsilon , \bx + \epsilon)$\\
\qquad\qquad  Augment $\bx$ with $\bx_{adv}$\\  
\textbf{return} $\btheta$

\caption{\footnotesize{ Model training with Prototype Conformity Loss.}}
\label{algo}
\end{algorithm}
\vspace{-1.0em}

\begin{table}[htp]
\centering
\caption{\small{Two network architectures: CNN-6 (MNIST, FMNIST) and ResNet-110 (CIFAR-10,100 and SVHN). Features are extracted in CNN-6 (after Layer 3 and two FC layers) and ResNet-110 (after Layer 3, 4 and FC layer) to impose the proposed $\mathcal{L}_{\text{PC}}$. Auxiliary branches are shown in green color.}} \vspace{-0.5em}
\label{table:models}

\resizebox{.67\width}{!}{
\begin{tabular}{c||c||c}
\hline 
\textbf{Layer \#}& \textbf{6-Conv Model} & \textbf{ResNet-110}\\
\hline \hline
\multirow{3}{*}{1} &
\multirow{3}{*}{\bigg[
  \begin{tabular}{c}
  Conv(32, $5\times 5$)\\
  PReLu($2\times 2$) 
  \end{tabular}
 \bigg] $\times 2 $}  & 
 
 \multirow{3}{*}{
  \begin{tabular}{c}
  Conv(16, $3\times 3$) + BN\\
  ReLU($2\times 2$)
  \end{tabular}
  } \\ 
& & \\
&& \\
\hline
\multirow{4}{*}{2}&
\multirow{4}{*}{\bigg[
  \begin{tabular}{c}
  Conv(64, $5\times 5$) \\
  PReLu($2\times 2$) 
  \end{tabular}
 \bigg] $\times 2 $}  &
 
 \multirow{4}{*}{\Bigg[
  \begin{tabular}{c}
  Conv(16, $1\times 1$) + BN\\
  Conv(16, $3\times 3$) + BN\\
  Conv(64, $1\times 1$) + BN\\
  \end{tabular}
 \Bigg] $\times 12 $}\\ 
&& \\
&& \\
&& \\
\hline
\multirow{5}{*}{3}&
\multirow{4}{*}{\bigg[
  \begin{tabular}{c}
  Conv(128, $5\times 5$) \\
  PReLu($2\times 2$) 
  \end{tabular}
 \bigg] $\times 2 $}  & 
 
  \multirow{4}{*}{\Bigg[
  \begin{tabular}{c}
  Conv(32, $1\times 1$) + BN\\
  Conv(32, $3\times 3$) + BN\\
  Conv(128, $1\times 1$) + BN\\
  \end{tabular}
 \Bigg] $\times 12 $}\\

&& \\
&& \\
& & \\
& \cellcolor{green!10} GAP {\color{red}$\rightarrow \mathcal{L}_{\text{PC}}$} & \cellcolor{green!10} (GAP$\rightarrow$FC(512) {\color{red}$\rightarrow \mathcal{L}_{\text{PC}}$})\\
\hline
\multirow{5}{*}{4} &

\multirow{5}{*}{FC(512) {\color{red}$\rightarrow \mathcal{L}_{\text{PC}}$}}  &
\multirow{4}{*}{\Bigg[
  \begin{tabular}{c}
  Conv(64, $1\times 1$) + BN\\
  Conv(64, $3\times 3$) + BN\\
  Conv(256, $1\times 1$)+ BN\\
  \end{tabular}
 \Bigg] $\times 12 $}\\
&&\\
&& \\
&& \\
&& \cellcolor{green!10} GAP {\color{red}$\rightarrow \mathcal{L}_{\text{PC}}$}\\
\hline
\multirow{2}{*}{5} &
\multirow{2}{*}{FC(64) {\color{red}$\rightarrow \mathcal{L}_{\text{PC}}$}}& \multirow{2}{*}{FC(1024) {\color{red}$\rightarrow \mathcal{L}_{\text{PC}}$}}\\
&\\
\hline
\multirow{2}{*}{6}&
\multirow{2}{*}{FC(10) {\color{blue}$\rightarrow \mathcal{L}_{\text{CE}}$}}& \multirow{2}{*}{FC(100/10) {\color{blue}$\rightarrow \mathcal{L}_{\text{CE}}$}}\\
&\\
\hline
\end{tabular}}
\vspace{-1em}
\end{table}

\begin{table*}[htp]
\caption{\small{Robustness of our model in \textit{white-box} and \textit{black-box} settings. Adversarial samples generated in the \textit{black-box} settings show negligible attack potential against our models. Here $\epsilon$ is the perturbation size and $c$ is the initial constant for C\&W attack. It can be seen that AdvTrain further complements the robustness of our models.}}
 \vspace{-0.75em}
\label{table:performance}

\begin{center}
  \vspace{-1em}
\resizebox{0.69\textwidth}{!}{
\begin{tabular}{cc|ccccc|ccccc}
\hline \hline
\multicolumn{1}{c}{}& \multicolumn{1}{c}{}  & \multicolumn{5}{c}{\textbf{White-Box Setting}} &  \multicolumn{5}{c}{\textbf{Black-Box Setting}}\\
\textbf{Training} &\textbf{No Attack} & \textbf{FGSM} & \textbf{BIM}  &  \textbf{C\&W}  & \textbf{MIM} 
 &  \textbf{PGD} & \textbf{FGSM} & \textbf{BIM}  &  \textbf{C\&W}  & \textbf{MIM} 
 &  \textbf{PGD} \\
\hline \hline
\multicolumn{12}{c}{MNIST ($\epsilon = 0.3 , c = 10$)} \\
\hline
Softmax & 98.71 & 4.9 & 0.0& 0.2 & 0.01 & 0.0 & 23.0 & 17.8 & 20.9 & 14.8 & 11.9\\
Ours & \textbf{99.53} & 31.1 & 23.3 & 29.1  & 24.7 & 19.9 & 78.3 & 72.7 & 77.2  & 74.5 & 69.5\\
Ours + AdvTrain$_{\tiny{FGSM}}$ & 99.44 & \textbf{53.1} & 36.6 & 40.9 & 37.0 & 34.5 & \textbf{85.6} & 81.0 & 82.3 & 81.4 & 78.2\\
Ours + AdvTrain$_{\tiny{PGD}}$ & 99.28 & 49.8 & \textbf{40.3} & \textbf{46.0} & \textbf{41.4}& \textbf{39.8} & 85.2 & \textbf{81.9} & \textbf{83.5} & \textbf{82.8} & \textbf{80.8}\\

\hline \hline

\multicolumn{12}{c}{CIFAR-10 ($\epsilon = 0.03 , c = 0.1$)} \\
\hline
Softmax & 90.80 & 21.4 & 0.0 &0.6  & 0.0 &0.01 & 39.0 & 30.1 & 31.8  & 30.9 & 29.1\\
Ours & 90.45 & 67.7 & 32.6 & 37.3 & 33.2  &27.2 & 85.5 & 83.7 & 83.3 & 81.9  & 76.4\\
Ours + AdvTrain$_{\tiny{FGSM}}$ & 91.28 & \textbf{75.8} & 45.9 & 45.7 & 44.7 & 42.5 & \textbf{88.9} & 87.6 & 87.4 & 88.2 & 84.5\\
Ours + AdvTrain$_{\tiny{PGD}}$ & \textbf{91.89} & 74.9 & \textbf{46.0} & \textbf{51.8} & \textbf{49.3} & \textbf{46.7} & 88.5& \textbf{88.3} & \textbf{88.2}& \textbf{88.5}& \textbf{
88.8} \\
\hline \hline

\multicolumn{12}{c}{CIFAR-100 ($\epsilon = 0.03 , c = 0.1$)} \\
\hline
Softmax & \textbf{72.65} &  20.0 & 4.2 & 1.1 & 3.52& 0.17 &  40.9 & 34.3 & 37.1 & 35.5& 30.7\\
Ours & 71.90 & 56.9 & 28.0& 31.1 & 28.7 & 25.9 & 65.3 & 64.5 & 64.1 & 64.8 & 62.8 \\
Ours + AdvTrain$_{\tiny{FGSM}}$ & 69.11& \textbf{61.3} & 32.3 & 35.2 & 33.3 & 31.4 & \textbf{66.1} & 65.2 & 65.7  & 65.5  & 63.4\\
Ours + AdvTrain$_{\tiny{PGD}}$ & 68.32 & 60.9 & \textbf{34.1} & \textbf{36.7}& \textbf{33.7} & \textbf{36.1} & 65.9 & \textbf{66.1} & \textbf{66.7} & \textbf{66.1}& \textbf{66.7}\\
\hline \hline

\multicolumn{12}{c}{F-MNIST ($\epsilon = 0.3 , c = 10$)} \\
\hline
Softmax & \textbf{91.51} & 8.7 & 0.1 & 0.2 & 0.0 & 0.0 & 46.7 & 29.3 & 30.8 & 29.5 & 26.0\\
Ours & 91.32 & 29.0 & 22.0 & 23.9 & 21.8 & 20.3 & 84.8 & 79.0 & 79.2 & 78.4 & 76.3\\
Ours + AdvTrain$_{\tiny{FGSM}}$ & 91.03 & \textbf{55.1} & 37.5 & 41.7 & 40.6 & 35.3 & \textbf{89.1} & 87.0 & 87.7 & 87.9 & 85.2\\
Ours + AdvTrain$_{\tiny{PGD}}$ & 91.30 & 47.2 & \textbf{40.1} & \textbf{44.6} & \textbf{41.3}& \textbf{40.7} & 88.2 & \textbf{88.0} & \textbf{88.2} & \textbf{88.3}& \textbf{89.7}\\
\hline \hline

\multicolumn{12}{c}{SVHN ($\epsilon = 0.03 , c = 0.1$)} \\
\hline
Softmax & 93.45 & 30.6 & 6.2 & 7.1 & 7.3 & 9.6 & 48.1 & 30.3 & 31.4 & 33.5 & 21.5\\
Ours & \textbf{94.36} & 69.3 & 37.1 & 39.2 & 41.0 & 33.7 & 77.4 & 73.1 & 76.4 & 74.0 & 70.1\\
Ours + AdvTrain$_{\tiny{FGSM}}$ & 94.18 & \textbf{80.1} & 47.4 & 51.9 & 45.6 & 40.5 & \textbf{90.1} & 87.4 & 88.0 & 87.6 & 84.4\\
Ours + AdvTrain$_{\tiny{PGD}}$ & \textbf{94.36} & 76.5 & \textbf{48.8} & \textbf{54.8} & \textbf{47.1} & \textbf{47.7} & 88.7 & \textbf{88.2} & \textbf{89.2} & \textbf{88.6} & \textbf{89.3}\\
\hline \hline
\vspace{-2em}
\end{tabular} 
}
\end{center}
\vspace{-1em}
\end{table*}

\subsection{Results and Analysis}

\noindent \textbf{\textit{White-Box} vs \textit{Black-Box} Settings:}
In an adversarial setting, there are two main threat models: \textit{white-box} attacks where the adversary possesses complete knowledge of the target model, including its parameters, architecture and the training method, and \textit{black-box attacks} where the adversary feeds perturbed images at test time (which are generated without any knowledge of the target model). We evaluate the robustness of our proposed defense against both \textit{white-box} and \textit{black-box} settings. Table~\ref{table:performance} shows our results for the different attacks described in Sec.~\ref{subsec:attacks}. The number of iterations for BIM, MIM and PGD are set to $10$ with a step size of $\epsilon/10$. The iteration steps for C\&W are $1,000$, with a learning rate of $0.01$. We report our model's robustness with and without adversarial training for standard perturbation size \ie $\epsilon = 0.3$ for F/MNIST and and $\epsilon = 0.03$ for the CIFAR-10/100 and SVHN datasets.

Recent literature has shown transferability amongst deep models \cite{tramer2017space,kurakin2016adversarial_2,43405}, where adversarial images are effective even for the models they were never generated on. An adversary can therefore exploit this characteristic of deep models and generate generic adversarial samples to attack unseen models. Defense against \textit{black-box} attacks is therefore highly desirable for secure deployment of machine learning models \cite{papernot2016limitations}. To demonstrate the effectiveness of our proposed defense under \textit{black-box} settings, we generate adversarial samples using a VGG-19 model, and feed them to the model trained using our proposed strategy. Results in Table~\ref{table:performance} show that \textit{black-box} settings have negligible attack potential against our model. For example, on the CIFAR-10 dataset, where our model's accuracy on clean images is $91.89\%$, even the strongest iterative attack (PGD-0.03) fails, and our defense retains an accuracy of $88.8\%$.

\noindent \textbf{Adversarial Training} has been shown to enhance many recently proposed defense methods \cite{kurakin2018adversarial}. We also evaluate the impact of adversarial training (AdvTrain) in conjunction with our proposed defense. For this, we jointly train our model on clean and attacked samples, which are generated using FGSM \cite{43405} and PGD \cite{madry2017towards} by uniformly sampling $\epsilon$ from an interval of [0.1, 0.5] for MNIST and F-MNIST and [0.01, 0.05] for CIFAR and SVHN. Results in Table~\ref{table:performance} indicate that AdvTrain further complements our method and provides an enhanced robustness under both \textit{black-box} and \textit{white-box} attack settings.

\noindent \textbf{Adaptive White-box Settings:} Since at inference time, our model performs conventional softmax prediction, we evaluated the robustness against standard \textit{white-box} attack settings, to be consistent with existing defenses. Now, we also experiment in an \textit{adaptive white-box} setting where the attack is performed on the joint PC+CE loss (with access to learned prototypes). Negligible performance drop is observed in adaptive settings (see Table~\ref{table:performance-total-loss}).

\begin{table}[htp] 
\caption{\small{Robustness in \textit{adaptive white-box} attack settings. The performance for conventional attacks (where CE is the adversarial loss) is shown in blue. $^{*}$ indicates adversarially trained models.}}
\label{table:performance-total-loss}
\begin{center}
\vspace{-1.5em}
\resizebox{\columnwidth}{!}{
\begin{tabular}{c|c|cccc}
\hline 
\cellcolor{black!10} \textbf{Training} &\cellcolor{black!10} \textbf{No Attack} &\cellcolor{black!10}  \textbf{FGSM} &\cellcolor{black!10}  \textbf{BIM} &\cellcolor{black!10}  \textbf{MIM}& \cellcolor{black!10}  \textbf{PGD} \\
\hline \hline

\multicolumn{6}{c}{CIFAR-10 ($\epsilon = 8/255$)} \\
\hline
Ours & 90.45 & 66.90 \textcolor{blue}{(67.7)} &31.29 \textcolor{blue}{(32.6)}& 32.84 \textcolor{blue}{(33.2)}& 27.09 \textcolor{blue}{(27.2)}\\
Ours$^*_{\text{\tiny{FGSM}}}$ & 91.28 & 74.24 \textcolor{blue}{(75.8)} & 44.05 \textcolor{blue}{(45.9)}& 43.77 \textcolor{blue}{(44.7)} & 41.32 \textcolor{blue}{(42.5)} \\
Ours$^*_{\text{\tiny{PGD}}}$ & 91.89 & 74.31 \textcolor{blue}{(74.9)}& 44.85 \textcolor{blue}{(46.0)}& 47.31 \textcolor{blue}{(49.3)}& 44.75 \textcolor{blue}{(46.7)}\\

\hline \hline

\multicolumn{6}{c}{F-MNIST ($\epsilon = 0.3/1$)} \\
\hline
Ours  & 91.32 & 28.1 \textcolor{blue}{(29.0)}& 21.7 \textcolor{blue}{(22.0)}& 20.3 \textcolor{blue}{(21.8)}& 19.5 \textcolor{blue}{(20.3)}\\
Ours$^*_{\text{\tiny{FGSM}}}$ & 91.03 & 53.3 \textcolor{blue}{(55.1)}& 36.0 \textcolor{blue}{(37.5)}& 39.3 \textcolor{blue}{(40.6)}& 34.7 \textcolor{blue}{(35.3)}\\
Ours$^*_{\text{\tiny{PGD}}}$ & 91.30 & 46.0 \textcolor{blue}{(47.2)}& 40.1 \textcolor{blue}{(40.1)}& 40.7 \textcolor{blue}{(41.3)}& 39.7 \textcolor{blue}{(40.7)}\\
\hline

\end{tabular} 
}
\vspace{-1.5em}
\end{center}

\end{table}

\subsection{Comparison with Existing Defenses}
\label{subsec:comparison}
\vspace{-0.5em}

We compare our method with recently proposed state-of-the art proactive defense mechanisms, which alter the network or use modified training loss functions. To this end, we compare with \cite{kurakin2016adversarial_2}, which injects adversarial examples into the training set and generates new samples at each iteration. We also compare with \cite{pang2019improving}, which introduces an Adaptive Diversity Promoting (ADP) regularizer to improve adversarial robustness. Further, we compare with an input gradient regularizer mechanism \cite{ross2018improving} that penalizes the degree to which input perturbations can change a model's predictions by regularizing the gradient of the cross-entropy loss. Finally, we compare with the current state-of-the-art Min-Max optimization based defense \cite{madry2017towards}, which augments the training data with adversarial examples, causing the maximum gradient increments to the loss within a specified $l_\infty$ norm.
The results in Tables~\ref{table-cifar10-comparison}~,~\ref{table-mnist-comparison}~and~\ref{table-cifar100-comparison} in terms of retained classification accuracy on different datasets show that our method significantly outperforms all existing defense schemes by a large margin. The performance gain is more pronounced for the strongest iterative attacks (\eg C\&W and PGD) with large perturbation budget $\epsilon$. For example, our method achieves a relative gain of $20.6\%$ (AdvTrain models) and $41.4\%$ (without AdvTrain) compared to the $2^{\text{nd}}$ best methods on the CIFAR-10 and MNIST datasets respectively for the PGD attack.
On CIFAR-100 dataset, for the strongest PGD attack with $\epsilon=0.01$, the proposed method achieves $38.9\%$ compared with $18.3\%$ by ADP \cite{pang2019improving}, which, to the best of our knowledge, is the only method in the literature evaluated on the CIFAR-100 dataset. Our results further indicate that adversarial training consistently compliments our method and augments its performance across all evaluated datasets.

\begin{table}[!htp]
\caption{\small{Comparison on \textbf{CIFAR-100} dataset for \textit{white-box} adversarial attacks (numbers shows robustness, higher is better). $^*$ sign denotes adversarially trained models. For our model, we report results without adversarial training (Ours) and with adversarially generated images from FGSM (Ours$^*_{f}$) and PGD (Ours$^*_{p}$ ) attacks.}} \vspace{-1.5em}
\label{table-cifar100-comparison}
\begin{center}
\resizebox{.47\textwidth}{!}{
\begin{tabular}{c||c|c|c||c|c|c}

\hline \hline
Attacks & Params. & Baseline & ADP \cite{pang2019improving} & Ours  & Ours$^*_{f}$  & Ours$^*_{p}$\\
\hline \hline
{No Attack}   & - &\cellcolor{yellow!40} \textbf{72.6} & 70.2 &\cellcolor{blue!10} 71.9  & 69.1& 68.3\\
\hline
\multirow{2}{*}{BIM}   & $\epsilon = 0.005 $ &21.6 &26.2 & 44.8&\cellcolor{blue!10} 55.1 &\cellcolor{yellow!40}  \textbf{55.7}\\
 & $\epsilon = 0.01 $ & 10.1 & 14.8 & 39.8 &\cellcolor{blue!10} 46.2 &\cellcolor{yellow!40} \textbf{46.9} \\
\hline
\multirow{2}{*}{MIM}   & $\epsilon = 0.005 $ & 24.2 & 29.4 &46.1 &\cellcolor{blue!10} 56.7 &\cellcolor{yellow!40} \textbf{57.1} \\
 & $\epsilon = 0.01 $ & 11.2 & 17.2& 40.6 &\cellcolor{blue!10} 43.8 &\cellcolor{yellow!40} \textbf{45.9} \\
\hline

\multirow{2}{*}{PGD}   & $\epsilon = 0.005 $ & 26.6 & 32.1 & 42.2 &\cellcolor{blue!10} 53.6 &\cellcolor{yellow!40} \textbf{55.0} \\
& $\epsilon = 0.01 $ &11.7 & 18.3 & 38.9 &\cellcolor{blue!10} 40.1 &\cellcolor{yellow!40}  \textbf{44.0}\\
\hline \hline
\end{tabular}}
\vspace{-1.5em}
\end{center}
\end{table}

Additionally we compare our model's performance with a close method proposed by Song \textit{et al.} \cite{song2018improving} in Table~\ref{table:comparison-iclr19}, where our approach outperforms them by a significant margin. Besides a clear improvement, we discuss our main distinguishing features below: \textbf{(a)} Our approach is based on a ``deeply-supervised" loss that prevents changes to the outputs within the limited perturbation budget. Such supervision paradigm is the main contributing factor towards our improved results (See Table~\ref{table-ablation}). \textbf{(b)} \cite{song2018improving} focuses on domain adaption between adversarial and natural samples without any constraint on the intermediate feature representations. In contrast, we explicitly enforce the hidden layer activations to be maximally separated in our network design. \textbf{(c)} \cite{song2018improving} only considers adversarially trained models, while we demonstrate clear improvements both with and without adversarial training (a more challenging setting). In Table~\ref{table:comparison-iclr19} we have followed the exact model settings used in \cite{song2018improving}.

\begin{SCtable}[][htp]
\hspace{-1em}
\caption{\small{Comparison of our approach with \cite{song2018improving} on 4 datasets. \colorbox{green!10}{Green} rows show results for \cite{song2018improving} and \colorbox{blue!10}{blue} for our models.}}
\label{table:comparison-iclr19}
\centering
\resizebox{0.30\textwidth}{!}{
\begin{tabular}{c||cccc}
\hline 
\textbf{Dataset} &\textbf{Clean} & \textbf{FGSM}  &\textbf{MIM} & \textbf{PGD}  \\
\hline \hline 

\multirow{2}{*}{F-MNIST ($\epsilon=0.1$)} &\cellcolor{green!10} 85.5 &\cellcolor{green!10} 78.2 &\cellcolor{green!10} 68.8 &\cellcolor{green!10} 68.6 \\
&\cellcolor{blue!10} \textbf{91.3} &\cellcolor{blue!10} \textbf{86.6} &\cellcolor{blue!10} \textbf{80.1} &\cellcolor{blue!10} \textbf{79.4} \\
\hline
\multirow{2}{*}{SVHN ($\epsilon=0.02$)} &\cellcolor{green!10} 82.9 &\cellcolor{green!10} 57.2 &\cellcolor{green!10} 53.9 &\cellcolor{green!10} 53.2 \\
&\cellcolor{blue!10} \textbf{94.4} &\cellcolor{blue!10} \textbf{87.1} &\cellcolor{blue!10} \textbf{82.2} &\cellcolor{blue!10} \textbf{80.7} \\
\hline
\multirow{2}{*}{CIFAR-10 ($\epsilon=4/255$)} &\cellcolor{green!10} 84.8 &\cellcolor{green!10} 60.7 &\cellcolor{green!10} 59.0 &\cellcolor{green!10} 58.1 \\
&\cellcolor{blue!10} \textbf{91.9} &\cellcolor{blue!10} \textbf{85.3} &\cellcolor{blue!10} \textbf{70.1} &\cellcolor{blue!10} \textbf{69.4} \\
\hline
\multirow{2}{*}{CIFAR-100 ($\epsilon=4/255$)} &\cellcolor{green!10} 61.6 &\cellcolor{green!10} 29.3 &\cellcolor{green!10} 27.3 &\cellcolor{green!10} 26.2 \\
&\cellcolor{blue!10} \textbf{71.9} &\cellcolor{blue!10} \textbf{49.1} &\cellcolor{blue!10} \textbf{40.7} &\cellcolor{blue!10} \textbf{38.6} \\
\hline
\end{tabular} 
}
\end{SCtable}


\vspace{-0.5em}
\subsection{Transferability Test}
\vspace{-0.50em}
We investigate the transferability of attacks on CIFAR-10 dataset between a standard VGG-19 model, adversarially trained VGG-19 \cite{kurakin2016adversarial_2}, Madry \textit{et al.}'s \cite{madry2017towards} and our model. We report the accuracy of target models (columns) on adversarial samples generated from source models (rows) in Table~\ref{table-transferability}. Our results yield the following findings: \vspace{1mm} \\ 
\textbf{Improved \textit{black-box} robustness:} As noted in \cite{athalye2018obfuscated}, a model that gives a false sense of security due to obfuscated gradients can be identified if the \textit{black-box} attacks are stronger than \textit{white-box}. In other words, robustness of such a model under \textit{white-box} settings is higher than under \textit{black-box} settings. It was shown in \cite{athalye2018obfuscated} that most of the existing defenses obfuscate gradients. Madry \emph{et al.}'s approach \cite{madry2017towards} was endorsed by \cite{athalye2018obfuscated} to not cause gradient masking. The comparison in Table~\ref{table-transferability} shows that our method outperforms \cite{madry2017towards}.\vspace{1mm}\\ 
\textbf{Similar architectures increase transferability:} Changing the source and target network architectures decreases the transferability of an attack. The same architectures (\eg VGG-19 and its AdvTrain counterpart, as in Table~\ref{table-transferability}) show increased robustness against \textit{black-box} attacks generated from each other.

\begin{table}[h]
\caption{ \small{Transferability Test on CIFAR-10: PGD adversaries are generated with $\epsilon = 0.03$, using the source network, and then evaluated on target model. Underline denotes robustness against \textit{white-box} attack. Note that adversarial samples generated on our model are highly transferable to other models as \textit{black-box} attacked images.}} \vspace{-1.5em}
\label{table-transferability}
\begin{center}
\resizebox{.45\textwidth}{!}{

\begin{tabular}{ c||c|c|c|c}
\hline \hline
\multirow{2}{*}{\diagbox{Source}{Target}} & \multirow{2}{*}{VGG-19} & \multirow{2}{*}{AdvTrain \cite{kurakin2016adversarial_2}} & \multirow{2}{*}{Madry \textit{et al.} \cite{madry2017towards}} & \multirow{2}{*}{Ours}   \\
&  &  & & \\
\hline \hline
VGG-19 & \underline{0.0} & 16.20 & 52.71 & 88.80\\
AdvTrain \cite{kurakin2016adversarial_2}  & 12.43 & \underline{0.0}& 49.80 & 72.53\\ 
Madry \textit{et al.} \cite{madry2017towards} & 58.91  & 67.32 & \underline{43.70} & 71.72\\  
Ours  & 50.31 & 61.02 & 66.70 & \underline{49.10}\\ 
\hline
\end{tabular} }
\vspace{-1.5em}
\end{center}
\end{table}

\subsection{Ablation Analysis}
\label{subsec:ablation}
\vspace{-0.25em}
\noindent \textbf{$\mathbf{\mathcal{L}_\text{PC}}$ at Different Layers:} We investigate the impact of our proposed prototype conformity loss ($\mathcal{L}_\text{PC}$) at different depths of the network. Specifically, as shown in Table~\ref{table-ablation}, we apply $\mathcal{L}_\text{PC}$ individually after each layer (see Table~\ref{table:models} for architectures) and in different combinations. We report the achieved results on the CIFAR-10 dataset for clean and perturbed samples (using FGSM and PGD attacks) in Table~\ref{table-ablation}. The network without any $\mathcal{L_\text{PC}}$ loss is equivalent to a standard softmax trained model. It achieves good performance on clean images, but fails under both \textit{white-box} and \textit{black-box} attacks (see Table~\ref{table:performance}). The models with $\mathcal{L}_\text{PC}$ loss at initial layers are unable to separate deep features class-wise, thereby resulting in inferior performance. Our proposed $\mathcal{L_\text{PC}}$ loss has maximum impact in the deeper layers of the network. This justifies our choice of different layers for $\mathcal{L}_\text{PC}$ loss, indicated in  Table~\ref{table:models}.

\begin{table}[h]
\caption{\small{\textbf{Ablation Analysis} with $\mathcal{L}_\text{PC}$ applied at different layers of ResNet-110 (Table~\ref{table:models}) for CIFAR-10 dataset.}}
\vspace{-1.5em}
\label{table-ablation}
\begin{center}
\resizebox{.34\textwidth}{!}{
\begin{tabular}{c||c|c|c}
\hline \hline
\multirow{2}{*}{Layer \#} & No Attack & FGSM & PGD \\
& $\epsilon = 0$ & $\epsilon = 0.03$ & $\epsilon = 0.03$\\
\hline \hline
 None & 90.80 & 21.40 & 0.01 \\
 Layer 1         &  74.30 & 23.71 & 0.01 \\ 
 Layer 2         &  81.92 & 30.96 & 8.04 \\  
 Layer 3         &  88.75& 33.74 & 10.47 \\ 
 Layer 4         &  90.51& 39.90&  11.90\\ 
 Layer 5         &  91.11& 47.02& 13.56\\  
 Layer 4+5     &  90.63& 55.36 &20.70   \\
\cellcolor{green!10} Layer 3+4+5 &  \cellcolor{green!10}90.45 & \cellcolor{green!10}67.71 &\cellcolor{green!10} 27.23  \\
\hline \hline
\end{tabular}
}
\end{center}
\vspace{-2em}
\end{table}

\begin{table*}[tp]
\caption{\small{Comparison on \textbf{CIFAR-10} dataset for \textit{white-box} adversarial attacks (numbers shows robustness, higher is better). $^*$ sign denotes adversarially trained models. For our model, we report results without adversarial training (Ours) and with adversarially generated images from FGSM (Ours$^*_{f}$) and PGD (Ours$^*_{p}$ ) attacks.}} \vspace{-1.5em}
\label{table-cifar10-comparison}
\begin{center}
\resizebox{.95\textwidth}{!}{
\begin{tabular}{c||c|c|c|c|c|c|c||c|c|c}

\hline \hline
Attacks & Params. & Baseline &AdvTrain \cite{kurakin2016adversarial_2}$^*$ &Yu \textit{et al.} \cite{yu2018interpreting}$^*$ &Ross \textit{et al.} \cite{ross2018improving}$^*$ &Pang \textit{et al.} \cite{pang2019improving}$^*$&Madry \textit{et al.} \cite{madry2017towards}$^*$ & Ours & Ours$^*_{f}$  & Ours$^*_{p}$\\
\hline \hline
{No Attack}   & - & 90.8  & 84.5 & 83.1 & 86.2 & 90.6 & 87.3 & 90.5 & \cellcolor{blue!10}91.3 & \cellcolor{yellow!40}\textbf{91.9}\\
\hline
\multirow{2}{*}{FGSM}   & $\epsilon = 0.02 $ & 36.5  &44.3 & 48.5& 39.5 & 61.7 & 71.6 & 72.5 &\cellcolor{yellow!40} \textbf{80.8} &\cellcolor{blue!10}  78.5\\
 & $\epsilon = 0.04 $ & 19.4 &31.0& 38.2 & 20.8 & 46.2 & 47.4& 56.3 &\cellcolor{yellow!40} \textbf{70.5}& \cellcolor{blue!10} 69.9\\
\hline
\multirow{2}{*}{BIM} & $\epsilon = 0.01 $ & 18.5 & 22.6 & 62.7 & 19.0 & 46.6 & 64.3 & 62.9 &\cellcolor{blue!10} 67.9 &\cellcolor{yellow!40} \textbf{74.5}\\
 & $\epsilon = 0.02 $ & 6.1 & 7.8& 39.3 & 6.9 & 31.0 & 49.3 &40.1 & \cellcolor{blue!10} 51.2&\cellcolor{yellow!40} \textbf{57.3}\\
\hline
\multirow{2}{*}{MIM}   & $\epsilon = 0.01 $ & 23.8 &23.9& - & 24.6 &  52.1 & 61.5  & 64.3 & \cellcolor{blue!10} 68.8&\cellcolor{yellow!40}\textbf{74.9}\\
 & $\epsilon = 0.02 $ & 7.4 & 9.3 & - & 9.5 & 35.9& 46.7 &  42.3 &\cellcolor{blue!10} 53.8&\cellcolor{yellow!40} \textbf{60.0}\\
\hline

\multirow{3}{*}{C\&W}   & $c = 0.001 $ & 61.3 & 67.7& 82.5 & 72.2 & 80.6 & 84.5 & 84.3 &\cellcolor{blue!10} 91.0&\cellcolor{yellow!40} \textbf{91.3}\\
 & $c = 0.01 $ & 35.2 & 40.9& 62.9 & 47.8 & 54.9& 65.7&  63.5 & \cellcolor{blue!10} 72.9&\cellcolor{yellow!40} \textbf{73.7}\\
 & $c = 0.1 $ & 0.6 & 25.4& 40.7 & 19.9 & 25.6 & 47.9 & 41.1 & \cellcolor{blue!10} 55.7&\cellcolor{yellow!40} \textbf{60.5}\\
\hline

\multirow{2}{*}{PGD}   & $\epsilon = 0.01 $ & 23.4 & 24.3 & - & 24.5 & 48.4 & 67.7 & 60.1  &\cellcolor{blue!10} 68.3&\cellcolor{yellow!40} \textbf{75.7}\\
& $\epsilon = 0.02 $ & 6.6 & 7.8& - & 8.5 &  30.4 & 48.5 & 39.3 &\cellcolor{blue!10} 50.6 & \cellcolor{yellow!40} \textbf{58.5}\\
\hline \hline

\end{tabular}}
\vspace{-1.5em}
\end{center}
\end{table*}

\begin{table*}[tp]

\caption{\small{Comparison on \textbf{MNIST} dataset for \textit{white-box} adversarial attacks (numbers shows robustness, higher is better). $^*$ sign denotes adversarially trained models. For our model, we report results without adversarial training (Ours) and with adversarially generated images from FGSM (Ours$^*_{f}$) and PGD (Ours$^*_{p}$ ) attacks.}} \vspace{-1.5em}
\label{table-mnist-comparison}
\begin{center}
\resizebox{.85\textwidth}{!}{
\begin{tabular}{c||c|c|c|c|c|c||c|c|c}
\hline \hline

Attacks &  Params. & Baseline & AdvTrain \cite{kurakin2016adversarial_2}$^*$ &Yu \textit{et al.} \cite{yu2018interpreting} &Ross \textit{et al.} \cite{ross2018improving} &Pang \textit{et al.} \cite{pang2019improving} & Ours & Ours$^*_{f}$  & Ours$^*_{p}$ \\
\hline \hline
{No Attack}   &  - & 98.7 & 99.1& 98.4& 99.2&\cellcolor{yellow!40} \textbf{99.5} &\cellcolor{yellow!40} \textbf{99.5} & \cellcolor{blue!10} 99.4& 99.3 \\
\hline
\multirow{2}{*}{FGSM}   & $\epsilon = 0.1 $ & 58.3 & 73.0  & 91.6& 91.6 & 96.3 &\cellcolor{blue!10} 97.1  &\cellcolor{yellow!40} \textbf{97.2} & 96.5\\
 & $\epsilon = 0.2 $ & 12.9 & 52.7 & 70.3 & 60.4 & 52.8 & 70.6 &\cellcolor{yellow!40} \textbf{80.0} &\cellcolor{blue!10} 77.9\\
\hline
\multirow{2}{*}{BIM} &   $\epsilon = 0.1 $ & 22.5 & 62.0 & 88.1 & 87.9 & 88.5 & 90.2&\cellcolor{blue!10} 92.0 & \cellcolor{yellow!40}\textbf{92.1}\\
 & $\epsilon = 0.15 $ & 12.2 & 18.7 &\cellcolor{blue!10} 77.1 & 32.1 & 73.6 & 76.3& 76.5 & \cellcolor{yellow!40}\textbf{77.3}\\
\hline
\multirow{2}{*}{MIM}   &   $\epsilon = 0.1 $ & 58.3 & 64.5 &-& 83.7 & 92.0 & 92.1 &\cellcolor{blue!10} 92.7 & \cellcolor{yellow!40} \textbf{93.0}\\
 &  $\epsilon = 0.15 $ & 16.1 & 28.8 &-& 29.3& 77.5& 77.7 &\cellcolor{blue!10} 80.2 &\cellcolor{yellow!40} \textbf{82.0}\\
\hline

\multirow{3}{*}{C\&W}   &  $c = 0.1 $ & 61.6 & 71.1 & 89.2 &88.1 & 97.3 &\cellcolor{yellow!40} \textbf{97.7}& 97.1 & \cellcolor{blue!10} 97.6\\
 &  $c = 1.0 $ & 30.6 & 39.2 & 79.1 & 75.3 & 78.1& 80.4&\cellcolor{blue!10} 87.3&\cellcolor{yellow!40} \textbf{91.2}\\
 &  $c = 10.0 $ & 0.2 & 17.0 & 37.6 & 20.0 & 23.8 & 29.1 & \cellcolor{blue!10} 39.7&\cellcolor{yellow!40} \textbf{46.0}\\
\hline

\multirow{2}{*}{PGD}   & $\epsilon = 0.1 $ & 50.7 & 62.7 &-& 77.0 & 82.8 & 83.6 &\cellcolor{blue!10} 93.7 &\cellcolor{yellow!40} \textbf{93.9}\\
&  $\epsilon = 0.15 $ & 6.3 & 31.9 &-& 44.2 & 41.0 & 62.5 &\cellcolor{blue!10} 78.8 & \cellcolor{yellow!40} \textbf{80.2}\\
\hline \hline

\end{tabular}}
\vspace{-1.95em}
\end{center}
\end{table*}

\subsection{Identifying Obfuscated Gradients}
Recently, Athalye \textit{et.al.}\ \cite{athalye2018obfuscated} were successful in breaking several defense mechanisms in the \textit{white-box} settings by identifying that they exhibit a false sense of security. They call this phenomenon \textit{gradient masking}. Below, we discuss how our defense mechanism does not cause gradient masking on the basis of characteristics defined in \cite{athalye2018obfuscated, gilmer2018motivating}. 

\begin{figure}
    \centering
   {\includegraphics[trim={0cm 9.6cm 13.3cm 0cm}, clip, width=1\linewidth]{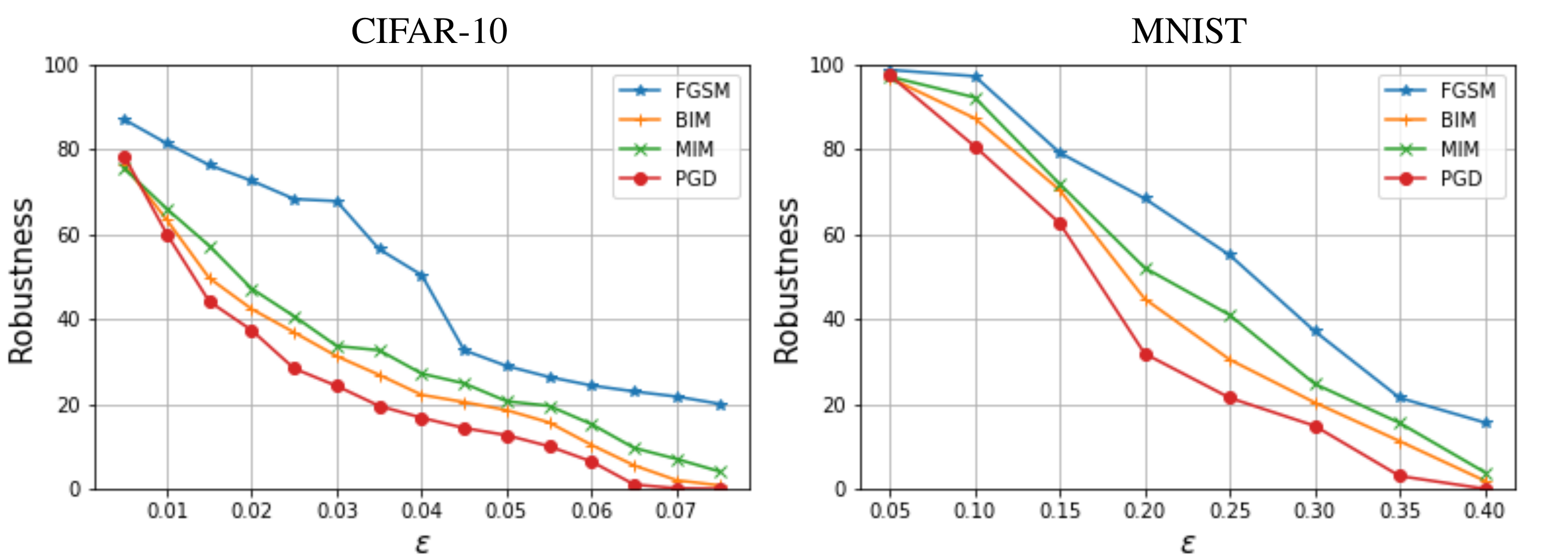} }
   \vspace{-2em}
    \caption{\small{Robustness of our model (without adversarial training) against  \textit{white-box} attacks for various perturbation budgets.}}
    \label{fig:epsilon}%
    \vspace{-0.5em}    
\end{figure}

\noindent
\textbf{Iterative attacks perform better than one-step attacks}: Our evaluations in Fig.~\ref{fig:epsilon} indicate that stronger iterative attacks (\eg BIM, MIM, PGD) in the \textit{white-box} settings are more successful at attacking the defense models than single-step attacks (FGSM in our case). 

\noindent
\textbf{Robustness against \textit{black-box} settings is higher than \textit{white-box} settings:} In \textit{white-box} settings, the adversary has complete knowledge of the model, so attacks should be more successful. In other words, if a defense does not suffer from obfuscated gradients, robustness of the model against \textit{white-box} settings should be inferior to that in the \textit{black-box} settings. Our extensive evaluations in Table~\ref{table:performance} show that the proposed defense follows this trend and therefore does not obfuscate gradients. 

\noindent
\textbf{Increasing the distortion bound ($\bm{\epsilon}$) decreases the robustness of defense}: On increasing the perturbation size, the success rate of the attack method should significantly increase monotonically. For an unbounded distortion, the classifier should exhibit 0\% robustness to the attack, which again is true in our case (see Fig.~\ref{fig:epsilon}).





\vspace{-0.35em}
\section{Conclusion}
\vspace{-0.5em}
Our findings provide evidence that the adversary's task is made difficult by incorporating a maximal separation constraint in the objective function of DNNs, which conventional cross-entropy loss fails to impose. Our theory and experiments indicate, if the adversarial polytopes for samples belonging to different classes are non-overlapping, the adversary cannot find a viable perturbation within the allowed budget. We extensively evaluate the proposed model against a diverse set of attacks (both single-step and iterative) in \textit{black-box} and \textit{white-box} settings and show that the proposed model maintains its high robustness in all cases. Through empirical evaluations, we further demonstrate that the achieved performance is not due to obfuscated gradients, thus the proposed model can provide significant security against adversarial vulnerabilities in deep networks. 

{\small
\bibliographystyle{ieee}
\bibliography{root}
}

\end{document}